%% file: main.tex
\documentclass{article}
\usepackage{ijcai21}

\usepackage{times}
\usepackage{soul}
\usepackage{url}
\usepackage[hidelinks]{hyperref}
\usepackage[utf8]{inputenc}
\usepackage[small]{caption}
\usepackage{graphicx}
\usepackage{amsmath}
\usepackage{amsthm}
\usepackage{booktabs}
\usepackage{algorithm}
\usepackage{algorithmic}
\usepackage{color}
\usepackage{CJKutf8} 
\usepackage{amssymb}

\newtheorem{example}{Example}

\newcommand{\newcite}[1]{\citeauthor{#1}~\shortcite{#1}}

\title{The Factual Inconsistency Problem in Abstractive Text Summarization: A Survey}

\author{
Yichong Huang\and
Xiachong Feng\and
Xiaocheng Feng\And
Bing Qin
\affiliations
Research Center for Social Computing and Information Retrieval\\
Harbin Institute of Technology, China\\
\emails
\{ychuang, xiachongfeng, xcfeng, qinb\}@ir.hit.edu.cn,
}

\begin{document}

\maketitle

\input{abstract}
\input{introduction}

\input{background}
\input{factual_consistency_evaluation_in_text_summarization}
\input{approaches_to_improve_factual_consistency_for_text_summarization_system}
\input{conclusion_and_future_directions}

\section{Acknowledgements}
Xiaocheng Feng is the corresponding author of this work. We thank the anonymous reviewers for their insightful comments. This work was supported by the National Key R\&D Program of China via grant 2020AAA0106502, National Natural Science Foundation of China (NSFC) via grant 62276078 and Natural Science Foundation of Heilongjiang via grant YQ2019F008.

\bibliographystyle{named}
\bibliography{main}

\end{document}

%% file: abstract.tex
\begin{abstract}

Recently, various neural encoder-decoder models pioneered by Seq2Seq framework have been proposed to achieve the goal of generating more abstractive summaries by learning to map input text to output text.
At a high level, such neural models can freely generate summaries without any constraint on the words or phrases used.
Moreover, their format is closer to human-edited summaries and output is more readable and fluent.
However, the neural model's abstraction ability is a double-edged sword. A commonly observed problem with the generated summaries is the distortion or fabrication of factual information in the article. 
This inconsistency between the original text and the summary has caused various concerns over its applicability, and the previous evaluation methods of text summarization are not suitable for this issue.
In response to the above problems, the current research direction is predominantly divided into two categories, one is to design fact-aware evaluation metrics to select outputs without factual inconsistency errors, and the other is to develop new summarization systems towards factual consistency.
In this survey, we focus on presenting a comprehensive review of these fact-specific evaluation methods and text summarization models.

\end{abstract}

%% file: introduction.tex
\section{Introduction}

Text summarization is one of the most important yet challenging tasks in Natural Language Processing (NLP) field.
It aims at condensing a piece of text to a shorter version that contains the main information from the original
document~\cite{mani1999advances,Nallapati2016AbstractiveTS}.

Text summarization approaches can be divided into two categories: \textit{extractive} and \textit{abstractive}. 

Extractive summarization is to find out the most salient sentences from the text by considering the statistical features and then arranging the extracted sentences to create the summary. 
Abstractive summarization, on the other hand, is a technique in which the summary is generated by generating novel sentences by either rephrasing or using the new words, instead of simply extracting the important sentences \cite{GUPTA201949}.
 
With the great strides of deep learning, neural-based abstractive summarization models are able to produce fluent and human-readable summaries~\cite{see-etal-2017-get}.
  
  \begin{figure}[!t]
  \centering
    \includegraphics[clip,width=0.9\columnwidth]{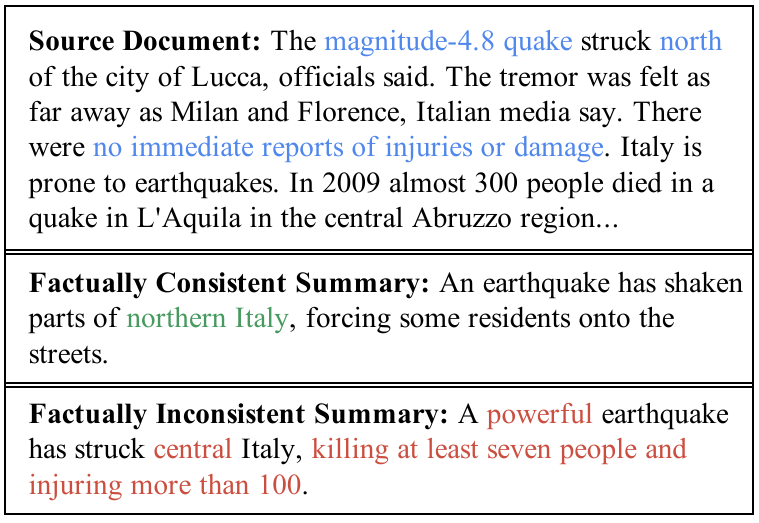}
    \caption{An example of factual inconsistency errors. The facts supported by the source document are marked in \textcolor[RGB]{71,154,95}{green}. Factual inconsistency errors are marked in \textcolor[RGB]{203,79,64}{red}.}
  \label{fig:error example}
\end{figure}

However, existing neural abstractive summarization models are highly prone to generate factual inconsistency errors. 
It refers to the phenomenon that the summary sometimes distorts or fabricates the facts in the article.
Recent studies show that up to 30\% of the summaries generated by abstractive models contain such factual
inconsistencies \cite{kryscinski-etal-2020-evaluating,falke-etal-2019-ranking}.
This brings serious problems to the credibility and usability of abstractive summarization systems. 
Figure 1 demonstrates an example article and excerpts of generated summaries.
 As shown, ``magnitude-4.8 earthquake" is exaggerated into a ``powerful quake," which will cause adverse social impacts.

\begin{figure*}[!h]
  \centering
    \includegraphics[clip,width=1.6\columnwidth]{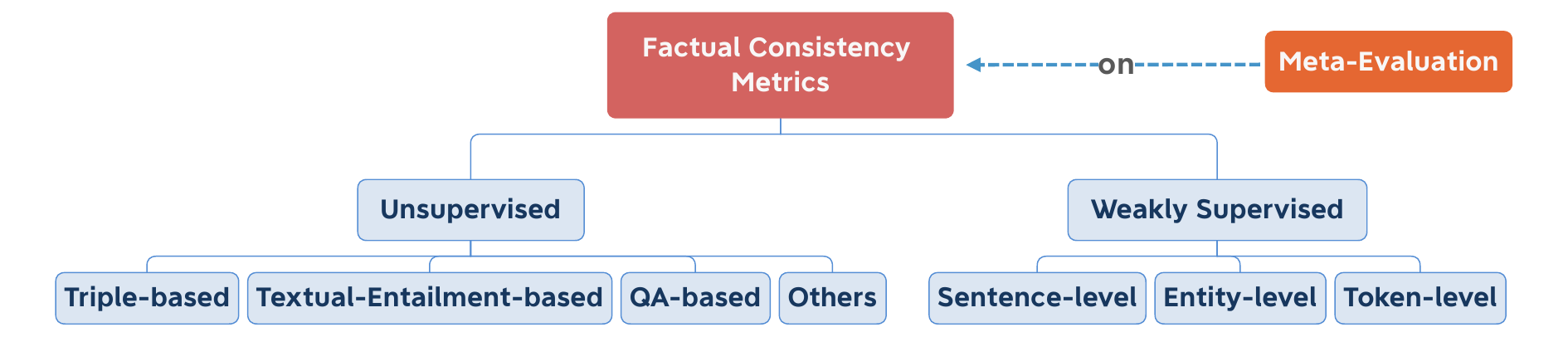}
    \caption{Factual consistency metrics.}
  \label{fig:factual consistency metrics}
\end{figure*}

 On the other hand, most existing summarization evaluation tools calculate N-gram overlaps between the generated summary and the human-written reference summary as the qualities of the generated summary, while neglecting fact-level consistency between two texts.
 \newcite{zhu2020boosting} point that the generated summaries are often high in token level
metrics like ROUGE \cite{lin-2004-rouge} but lack factual correctness.
For instance, the sentences ``I am having vacation in Hawaii." and ``I am not having vacation in Hawaii." share nearly all unigrams and bigrams despite having the opposite meaning.

To address the factual inconsistency issue, a lot of automatic factual consistency evaluation metrics and meta-evaluation for these metrics are proposed (\S\ref{section:factual consistency evaluation}). 
Besides, much effort has been devoted to optimizing factual consistency for summarization systems (\S\ref{section:factual consistency optimization}). 
In the past three years, more than twenty studies on factual consistency of summarization have been proposed.
Considering the large amount of effort dedicated to resolving the \textit{factual inconsistency} problem and the large interest of the NLP community in abstractive summarization, in this work we first introduce readers to the field of abstractive summarization. 
Then we focus on the \textit{factual inconsistency} problem by giving a broad overview of factual consistency evaluation and factual consistency optimization methods. 
Throughout this survey, we outline the lessons learned from the introduced methods and provide views on possible future directions.

%% file: background.tex
\section{Background}

In this section, we first introduce readers to abstractive summarization methods to better understand the reasons for generating factual inconsistency errors. 
Then we give the definition of the factual inconsistency error and the corresponding category.

\subsection{Abstractive Summarization methods}

Conventional abstractive summarization methods usually extract some keywords from the source document, and then reorder and perform linguistically motivated transformations to these keywords~\cite{13730}. 
However, previous paraphrase-based generation methods are easy to produce influent sentences~\cite{881692}. 

\newcite{Nallapati2016AbstractiveTS} first proposed to use an RNN (i.e. \textbf{encoder}) to encode the source document into a sequence of word vectors, and use another RNN (i.e. \textbf{decoder}) to generate a sequence of words as the generated summary based on the word vectors from encoder. 
The encoder and decoder could also be implemented by CNN~\cite{narayan-etal-2018-dont} and Transformer~\cite{vaswani2017attention}. 
The decoder of those sequence-to-sequence based neural text generation is a conditional language model, which  makes generating readable and fluent text possible~\cite{fan-etal-2018-hierarchical}.
However, most summarization systems are trained to maximize the log-likelihood of the reference summary at the word-level, which does not necessarily reward models for being faithful~\cite{maynez-etal-2020-faithfulness}.

\subsection{Factual Inconsistency Error} \label{subsection:facutal consistency error}

Factual inconsistency errors, i.e., facts inconsistent with the source document, could be divided into two categories:

\textbf{Intrinsic Error:} the fact that is \textbf{contradicted} to the source document, which is also referred to as ``intrinsic hallucination" in~\newcite{maynez-etal-2020-faithfulness}. In Figure~\ref{fig:error example}, the word ``central", which is contradicted to ``north" in the source document, belongs to this case.

\textbf{Extrinsic Error:} the fact that is \textbf{neutral} to the source document (i.e., the content that is neither supported nor contradicted by the source document), which is also referred to as ``extrinsic hallucination". As shown in Figure~\ref{fig:error example}, ``killing at least seven people and injuring more than 100", which is not reported in the source document, belongs to this case. 

It is worth mentioning that, existing factual consistency optimization methods mainly focus on intrinsic errors, and these two categories are not distinguished in factual consistency evaluation metrics.

\begin{table*}[t!]
\centering
\begin{tabular}{lrrrr}
\toprule
Metric  & Category & Summarization Dataset & Code  \\
\midrule
\newcite{10.1145/3292500.3330955}           & Triple-based              & Wikipedia     & \\ 

\newcite{falke-etal-2019-ranking}           & Textual-Entailment-based  & CNN/DM       & \\

\newcite{mishra2020looking}                 & Textual-Entailment-based  & CNN/DM       & \\

QAGS~\cite{wang-etal-2020-asking}             & QA-based                  & CNN/DM, XSum & 
\href{https://github.com/W4ngatang/qags}{\checkmark} \\

FEQA~\cite{durmus-etal-2020-feqa}             & QA-based                  & CNN/DM, XSum &
\href{https://github.com/esdurmus/feqa}{\checkmark}  \\

FactCC~\cite{kryscinski-etal-2020-evaluating}   & Weakly-Supervised-based   & CNN/DM       &
\href{https://github.com/salesforce/factCC}{\checkmark} \\

HERMAN~\cite{zhao-etal-2020-reducing}        & Weakly-Supervised-based   & XSum & \\

\newcite{zhou2020detecting}     & Weakly-Supervised-based   & XSum  &  \\
\bottomrule
\end{tabular}
\caption{List of factual consistency metrics. Code refers whether the code is available. \checkmark links to corresponding resource location.}
\label{tab:list of factual consistency metrics}
\end{table*}

%% file: factual_consistency_evaluation_in_text_summarization.tex
\section{Factual Consistency Metrics} \label{section:factual consistency evaluation}

We take the stock of factual consistency metrics, and then divide them into two categories: unsupervised and weakly supervised, as shown in Figure~\ref{fig:factual consistency metrics}. \textbf{Unsupervised metrics} use existing tools to evaluate factual consistency of summaries. According to tools that unsupervised metrics base on, we further divide unsupervised metrics into 4 types: Triple-based, Textual entailment-based, QA-based and Others. \textbf{Weakly supervised metrics} need to train on the factual consistency data, which consists of documents and model-generated summaries and factual consistency scores for each summaries. To compare factual consistency metrics with each other, \textbf{Meta-evaluations} for factual consistency rise up. We introduce 2 meta-evaluation works about factual consistency. Besides, we organize existing metrics into Table~\ref{tab:list of factual consistency metrics}.

\subsection{Unsupervised Metrics}

  \begin{figure}[h!]
  \centering
    \includegraphics[clip,width=0.9\columnwidth]{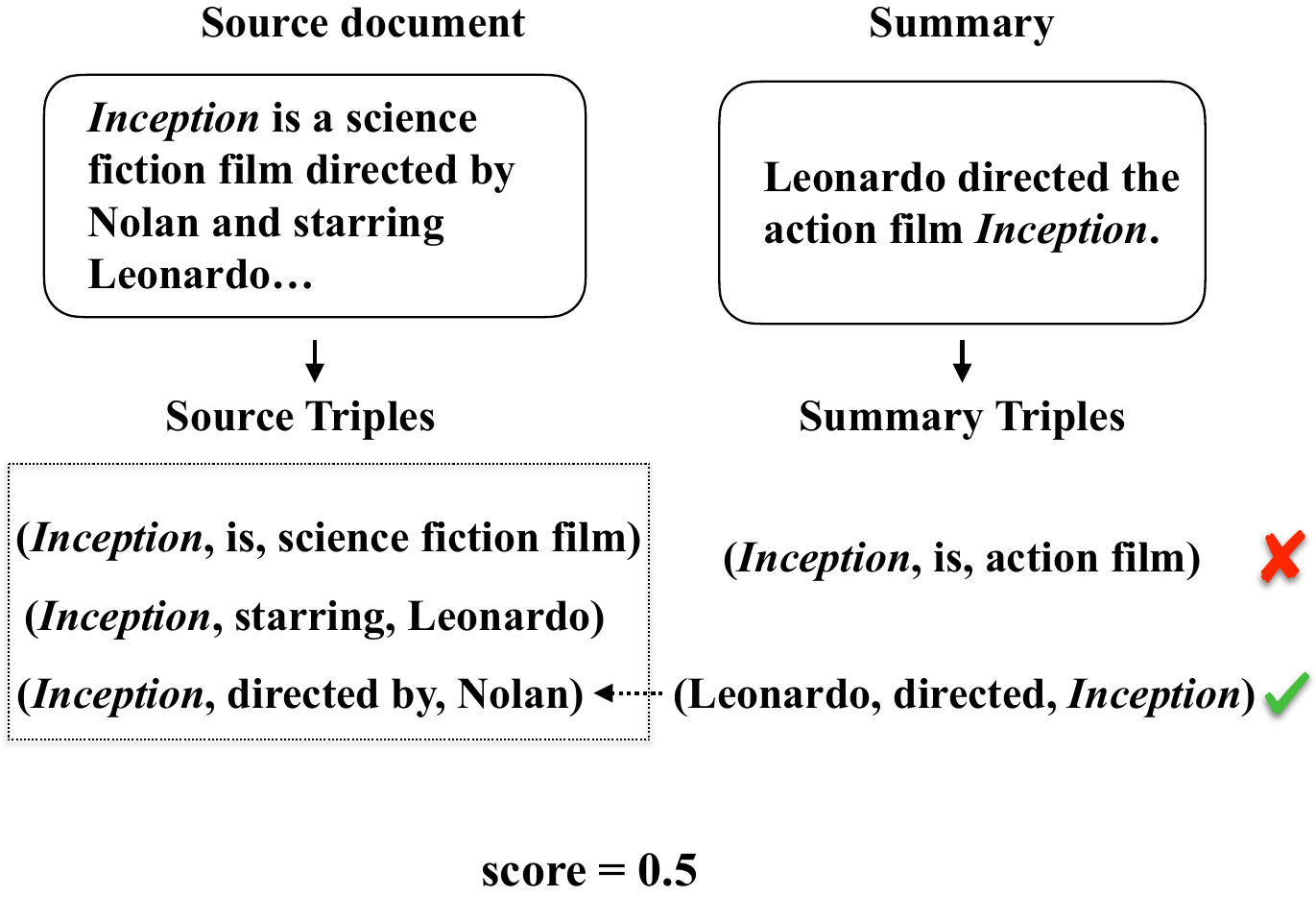}
    \caption{Triple-based factual consistency metrics.}
  \label{fig:triple-based factual consistency metrics}
\end{figure}

\subsubsection{Triple-based} \label{subsection:triple-based}
The most intuitive way to evaluating factual consistency is to count the fact overlap between generated summary and the source document, as shown in Figure~\ref{fig:triple-based factual consistency metrics}. 
Facts are usually represented by relation triples (\textit{subject, relation, object}), where the subject has a relation to the object. 
To extract triples, ~\newcite{10.1145/3292500.3330955} first try to use OpenIE tool~\cite{10.5555/1625275.1625705}. 
However, OpenIE extracts triples with an unspecified schema instead of a fixed schema. In unspecific schema extraction, relation is extracted from the text between subject and object. In fixed schema extraction, a relation is predicted from a pre-defined relations set, which could be viewed as a classification task. Unspecific schema extraction makes the extracted triples hard to compare with each other. As shown in Example~\ref{example:relation extraction with unspecified schema}, from two sentences expressing the same fact, we will get different triples that mismatch each other.
\begin{example}[Relation Extraction with Unspecific Schema] \label{example:relation extraction with unspecified schema}

\textbf{source document:} ``\textit{Obama was born in Hawaii}" $\Rightarrow$ \textit{(Obama, born in, Hawaii)}
\\
\textbf{summary:} ``\textit{Hawaii is the birthplace of Obama}" $\Rightarrow$ \textit{(Hawaii, is the birthplace of, Obama)}.
\end{example}

To resolve this problem, ~\newcite{10.1145/3292500.3330955} change to use relation extraction tools with fixed schema. Considering still the two sentences in Example~\ref{example:relation extraction with unspecified schema}, whether extracting from the source document or the summary, the extracted triples are \textit{(Hawaii, is the birthplace of, Obama)} in fixed schema extraction. This helps extracted triples easier to compare. 

\subsubsection{Textual-Entailment-based} \label{subsection:textual-entailment-based}

Following the idea that \textbf{a factually consistent summary is semantically entailed by the source document}, ~\newcite{falke-etal-2019-ranking} propose to use textual entailment prediction tools to evaluate the factual consistency for a summary. Textual entailment prediction, also known as Natural Language Inference (NLI), aims at detecting whether a text P (\textit{Premise}) could entail another text H (\textit{Hypothesis}). 
However, out-of-the-box entailment models do not yet offer the desired performance for factual consistency evaluation in text summarization. 
One reason is that the domain shift from the NLI dataset to the summarization dataset. 
Another one is that NLI models tend to rely on heuristics such as lexical overlap to explain the high entailment probability. 
As a consequence, existing NLI models generalize poorly in downstream tasks. 

To make NLI models more generalizable, \newcite{mishra2020looking} first conjecture that a key difference between the NLI datasets and downstream tasks concerns the length of the premise. Specifically, most existing NLI datasets consider one or at most a few sentences as the premise. However, most downstream NLP tasks such as text summarization and question answering consider the longer text as the premise, which requires reasoning over longer text. Reasoning over longer text needs a multitude of additional abilities like coreference resolution and abductive reasoning. To bridge this gap, they next create new long premise NLI datasets out of existing QA datasets for training a truly generalizable NLI model. After training on this new NLI dataset, the model gains significant improvement in the factual consistency evaluation task.

\subsubsection{QA-based} \label{subsection:qa-based}
\begin{figure}[t!]
  \centering
    \includegraphics[clip,width=0.9\columnwidth]{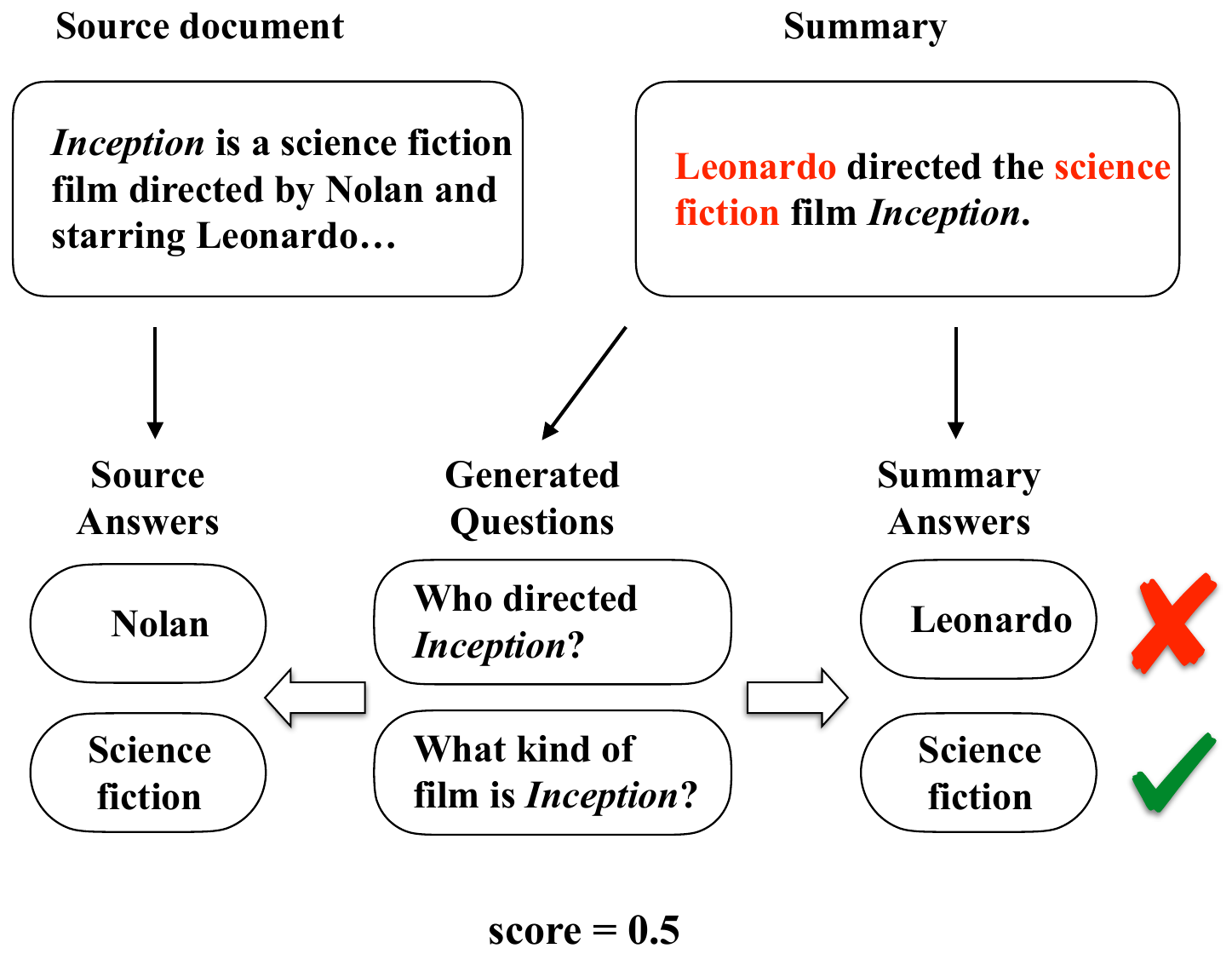}
    \caption{QA-based factual consistency metrics.}
  \label{fig:qa-based factual consistency metrics}
\end{figure}

Inspired by other question answering (QA) based automatic metrics in text summarization, \newcite{wang-etal-2020-asking,durmus-etal-2020-feqa} propose QA based factual consistency evaluation metrics QAGS and FEQA separately. These two metrics are all based on the intuition that if we ask questions about a summary and its source document, we will receive similar answers if the summary is factually consistent with the source document. As illustrated in Figure~\ref{fig:qa-based factual consistency metrics}, they are all consist of three steps: (1) Given a generated summary, a question generation (QG) model generates a set of questions about the summary, standard answers of which are named entities and key phrases in the summary. (2) Then using question answering (QA) model to answers these questions given the source document. (3) A factual consistency score is computed based on the similarity of corresponding answers. Because evaluating factual consistency at entity-level, these methods are more interpretable than textual-entailment-based methods. The reading comprehension ability of QG and QA models brings these methods promising performance in this task. However, these approaches are computationally expensive.

\subsubsection{Other Methods}
Besides the above methods specially designed, there are also several simple but effective methods to evaluate factual consistency, which are usually used as baselines. 
~\newcite{durmus-etal-2020-feqa} propose that a straightforward metric for factual consistency is the word overlap or semantic similarity between the summary sentence and the source document. The word overlap-based metrics compute ROUGE~\cite{lin-2004-rouge}, BLEU~\cite{papineni-etal-2002-bleu}, between the output summary sentence and each source sentence. And then taking the average score or maximum score across all the source sentences. The semantic similarity-based metric is similar to word overlap-based methods. Instead of using ROUGE or BLEU, this method uses BERTScore~\cite{Zhang*2020BERTScore:}. 
These two types of methods show a baseline level of effectiveness. And experiments in ~\newcite{durmus-etal-2020-feqa} show that word overlap-based methods work better in lowly abstractive summarization datasets like CNN/DM~\cite{hermann2015teaching}, semantic similarity-based method works better in highly abstractive summarization datasets like XSum~\cite{narayan-etal-2018-dont}.
Abstractiveness of the summarization dataset means the extent how abstract the reference summaries are against the source documents. Extremely, the summarization dataset is the least abstractive if all the reference summaries of which are directly extracted from the source document.

\subsection{Weakly Supervised Metrics} \label{subsection:model-based evaluation metrics}

Weakly supervised metrics design models specially for evaluating factual consistency. And these models are trained on synthetic data that are generated from the summarization dataset automatically, which side-steps the scarcity of the training data. According to objects to evaluated, existing metrics are divided into three categories: sentence-level, entity-level, and token-level.

\subsubsection{Sentence-level}
\newcite{kryscinski-etal-2020-evaluating} propose FactCC, a model to verify the factual consistency for a summary sentence given its source document. FactCC is implemented by fine-tuning pre-trained language model BERT~\cite{devlin-etal-2019-bert} as a binary classifier. And they further propose to automatically generate synthetic training data from the summarization dataset CNN/DM~\cite{hermann2015teaching}. Training examples are created by first sampling single sentences, later referred to as \textit{claims}, from the source documents. \textit{Claims} then pass through a set of textual transformations that output novel sentences with pseudo positive or negative labels. The positive examples are obtained through semantically invariant transformations like paraphrasing. The negative examples are obtained through semantically variant transformations like sentence negation and entity swap. In the test stage, FactCC takes the source document and a summary sentence as input and outputs the factual consistency label for this summary sentence. By simulating different kinds of factual inconsistency errors, this method gains some performance improvement in factual consistency evaluation. But at the same time, this rule-based dataset construction method brings a performance bottleneck. Because it is hard to simulate all types of factual inconsistency errors.

\subsubsection{Entity-level}
~\newcite{zhao-etal-2020-reducing} propose HERMAN, which focuses on evaluating factual consistency of quantity entities (e.g. numbers, dates, etc). HERMAN bases on sequence labeling architecture, in which input is the source document and summary, the output is a sequence of labels indicating which tokens consist of factual inconsistent quantity entities. The synthetic training data for HERMAN is automatically generated from the summarization dataset XSum~\cite{narayan-etal-2018-dont}. Rather than sampling document sentences as \textit{claims}, ~\newcite{zhao-etal-2020-reducing} use reference summary as \textit{claims}. And these \textit{claims} are directly labeled as positive summaries. The negative summaries are obtained by replacing the quantity entities in positive summaries. 

\subsubsection{Token-level}
~\newcite{zhou2020detecting} propose to evaluate factual consistency on token-level, which is more fine-grained and more explainable than sentence-level and entity-level evaluation. This token-level metric is implemented by fine-tuning pre-trained language model. Like ~\newcite{zhao-etal-2020-reducing}, reference summaries are also directly labeled as positive examples, and the negative examples are obtained by reconstructing part of reference summaries. This method shows higher correlations with human factual consistency evaluation.

These weakly supervised metrics attract much attention recently. But they still needs a large amount of human-annotated data (the source documents, model-generated summaries, and the factual consistency label for each summary) to achieve higher performance. However, such data are exceedingly expensive to produce in terms of both money and time. While existing weakly supervised methods automatically generate training data with heuristic, they use either document sentences or reference summaries to construct positive and negative examples. Nevertheless, both of them are different from summaries generated by summarization models. So although existing model-based methods show effectiveness in the training dataset, when applied in the real factual consistency evaluation scenario, the effect is very limited.

\begin{figure*}[!t]
  \centering
    \includegraphics[clip,width=1.6\columnwidth]{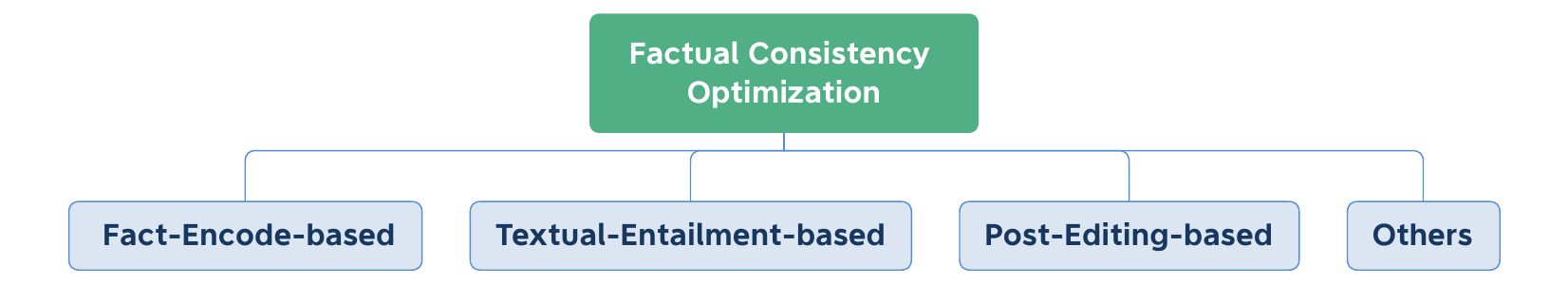}
    \caption{Factual consistency optimization methods.}
  \label{fig: factual consistency optimization}
\end{figure*}

\subsection{Meta Evaluation}

To verify the effectiveness of the above factual consistency metrics, most related works usually report the correlation between their own metric and human-annotated factual consistency score. However, it is still hard to compare each metric by the correlations as the diversity of annotating settings in different works and the disagreement among different annotator groups. To measure the effectiveness of different kinds of factual consistency metrics, ~\newcite{gabriel2020figure,koto2020ffci} conduct meta-evaluations of factual consistency in summarization. They evaluate the quality of several factual consistency metrics by computing the correlation between scores given by these metrics and scores measured by the same group of annotators. 

Through meta-evaluation, ~\newcite{koto2020ffci} find that \textbf{the semantic similarity-based method could reach state-of-the-art performance for factual consistency evaluation by searching optimal model parameters} (i.e. model layers of pre-trained language model in BERTScore) in highly abstractive summarization dataset XSum~\cite{narayan-etal-2018-dont}. Even so, the correlation with human evaluation is not more than 0.5. \textbf{Therefore, factual consistency evaluation is still an open issue in exploration.}

%% file: approaches_to_improve_factual_consistency_for_text_summarization_system.tex
\section{Factual Consistency Optimization} \label{section:factual consistency optimization}

\begin{table*}
\centering
\begin{tabular}{lrrr}
\toprule
Model  & Category & Summarization Dataset & Code \\
\midrule
FTSum~\cite{Cao:18}                     & Fact-Encode-based  & Gigaword     & \\

FASum~\cite{zhu2020boosting}            & Fact-Encode-based  & CNN/DM, XSum  & \\

ASGARD~\cite{huang-etal-2020-knowledge} & Fact-Encode-based  & CNN/DM, NYT   &
\href{https://github.com/luyang-huang96/GraphAugmentedSum}{\checkmark}\\

\newcite{gunel2019mind}                 & Fact-Encode-based  & CNN/DM       &   \\

\newcite{li-etal-2018-ensure}           & Textual-Entailment-based  & Gigaword &
\href{https://github.com/hrlinlp/entail_sum}{\checkmark} \\

SpanFact~\cite{dong-etal-2020-multi-fact} & Post-Editing-based  & CNN/DM, XSum, Gigaword    & \\

\newcite{cao-etal-2020-factual}         & Post-Editing-based    & CNN/DM    & \\

\newcite{matsumaru-etal-2020-improving} & Other                 & Gigaword, JAMUL   & \\

\newcite{mao2020constrained}            & Other                 & CNN/DM, XSum  & \\

\newcite{zhang-etal-2020-optimizing}    & Other                 & Radiology Reports Summarization & 
\href{https://github.com/yuhaozhang/summarize-radiology-findings}{\checkmark} \\

\newcite{yuan-etal-2020-faithfulness}   & Other                 & E-commerce Product Summarization & 
\href{https://github.com/ypnlp/coling}{\checkmark} \\
\bottomrule
\end{tabular}
\caption{List of factual consistency optimization methods. \checkmark links to corresponding resource location.}
\label{tab:list of factual consistency optimization methods}
\end{table*}

In this section, we provide an overview of approaches to optimizing summarization systems towards factual consistency. As illustrated in Figure~\ref{fig: factual consistency optimization}, we divide existing methods roughly into 4 classes according to principles that each method bases on: fact encode-based, textual entailment-based, post-editing-based, and other methods. Besides, we organize these methods into Table~\ref{tab:list of factual consistency optimization methods}.

\subsection{Fact-Encode-based}
In the earliest research about factual consistency, most works mainly focus on intrinsic factual inconsistency errors, i.e., the fact that is inconsistent with the source document. Intrinsic factual inconsistency errors convey wrongly the fact of the source document, which usually manifests as cross-combinations of the semantic units in different facts. For example, ``Jenny likes dancing. Bob likes playing football." $\Rightarrow$ ``Jenny likes playing football". It is the root cause for intrinsic errors that models misunderstand facts in the source document.

To help summarization systems understand correctly the facts, the most intuitive method is to explicitly model the facts in the source document, to augment the representation of facts. Following this idea, fact encode-based methods first extract facts in the source document, which are usually represented by relation triples consisting of (subject; relation; object). Then, these methods additionally encode the extracted facts into summarization models. According to the ways to encoding facts, these methods are divided into two categories: sequential encode and graph-based encode.

\subsubsection{sequential fact encode}
\newcite{Cao:18} introduce FTSum, which consists of two RNN encoders and one RNN decoder. FTSum concatenates the facts in the source document into a string which is named \textit{fact description}. One encoder encodes the source document and another encoder encodes the \textit{fact description}. The decoder attends the outputs from these two encoders when generating the summary. Even though experimental results show that FTSum reduces significantly factual inconsistent errors, it is hard for FTSum to capture the interactions between entities in all facts.

\subsubsection{graph-based fact encode}
To resolve this issue, ~\newcite{zhu2020boosting,huang-etal-2020-knowledge} propose to model the facts in the source document with knowledge graphs. FASum (Fact-Aware Summarization)~\cite{zhu2020boosting}, a transformer-based summarization model, uses a graph neural network (GNN) to learn the representation of each node (i.e., entities and relations) and fuses them into the summarization model. Comparing with FASum, ASGARD (Abstractive Summarization with Graph Augmentation and semantic-driven RewarD) ~\newcite{huang-etal-2020-knowledge} further uses multiple choice cloze reward to drive the model to acquire semantic understanding over the input.

In addition to enhancing the representation of facts in the source document, incorporating commonsense knowledge is also useful to facilitate summarization systems understanding the facts. Therefore, ~\newcite{gunel2019mind} sample relation triples from Wikidata to construct a commonsense knowledge graph. In this knowledge graph, TransE~\cite{10.5555/2999792.2999923}, the popular multi-relational data modeling method, is used to learn entity embeddings. And the summarization system attends to the embedding of related entities when encoding the source document. In this way, commonsense knowledge is incorporated into the summarization systems.

\subsection{Textual-Entailment-based}

Following the idea that a factual consistent summary is semantically entailed by the source document, ~\cite{li-etal-2018-ensure} propose an entailment-aware summarization model, which aims at incorporating entailment knowledge into the summarization model. Specifically, they propose a pair of entailment-aware encoder and decoder. The entailment-aware encoder is used to learn simultaneously summary generation and textual entailment prediction. And the entailment-aware decoder is implemented by entailment Reward Augmented Maximum Likelihood (RAML) training. RAML~\cite{norouzi2016reward} provides a computationally efficient approach to optimizing task-specific reward (loss) directly. In this model, the reward is the entailment score of generated summary.

\subsection{Post-Editing-based}

The above two kinds of methods optimize summarization systems towards factual consistency by modifying model structures. Different from those methods, post-editing-based methods enhance factual consistency of the final summary by post-editing the model-generated summaries which are viewed as \textbf{draft summaries} using fact corrects. Fact correctors take the source document and draft summary as input and generates the corrected summary as the final summary. 

Inspired by the QA span selection task, ~\newcite{dong-etal-2020-multi-fact} propose SpanFact, a suite of two span selection-based fact correctors, which corrects the entities in the draft summary in an iterative or auto-regressive manner respectively. Before performing fact correcting, one or all entities (one in the iterative manner, all in the auto-regressive manner) will be masked. Then SpanFact selects spans in the source document to replace corresponding mask tokens based on the understanding of the source document. 

To train SpanFact, ~\cite{dong-etal-2020-multi-fact} construct the dataset automatically. 

Human evaluation results shown that SpanFact successfully corrects about 26\% factually inconsistent summaries and wrongly corrupts less than 1\% factually consistent summaries. However, SpanFact is limited to correct entity errors.

Simpler than SpanFact, ~\newcite{cao-etal-2020-factual} propose an End-to-End fact corrector, which can correct more types of errors. This End-to-End fact corrector is implemented by fine-tuning pre-trained language model BART~\cite{lewis-etal-2020-bart} with artificial data. It takes and corrupted summary as input. The output is the corrected summary. Even though this method could correct more factually inconsistent errors than SpanFact conceptually, it has not outperform SpanFact in human evaluation result.

Both of ~\newcite{dong-etal-2020-multi-fact} and ~\newcite{cao-etal-2020-factual} choose to construct artificial training data automatically instead of using expensive human annotation. 

However, the gap between the training stage (which learns to correct the corrupted reference summaries) and the testing stage (which aims to correct the model-generated summaries) limits the performance of post-editing-based methods for the reason that corrupted reference summaries have a different data distribution with the model-generated summaries, which is the same as weakly supervised factual consistency metrics (\S\ref{subsection:model-based evaluation metrics}).

\subsection{Other}

Apart from the above methods, there are several simple but useful methods and domain-specific methods. 

~\newcite{matsumaru-etal-2020-improving} conjecture that one of the reasons why the model sometimes generates factually inconsistent summaries lies in unfaithful document-summary pairs, which are used for training the model. To mitigate this issue, they further propose to filter inconsistent training examples with a textual entailment classifier.

~\newcite{mao2020constrained} propose to improve factual consistency by applying constraints during the inference stage (i.e., beam search stage). Specifically, summarization models could end decoding only when all the constraints are met. And the constraints are important entities and keyphrases. Because this method only works at the inference stage as a plug-and-play module, it could be integrated into any abstractive summarization model without modifying its internal structure. However, how much improvement of factual consistency could be achieved by this method and how to design more useful constraints are still questions to explore.

Comparing to relatively open domain summarization, such as news field, optimization approaches towards factual consistency in special field are more different for their field characters. 
In the medical field, \newcite{zhang-etal-2020-optimizing} propose to optimize the factual consistency of radiology reports summarization. \newcite{shah2021nutribullets} propose to optimize the factual consistency of health and nutrition summarization.

In e-commerce field, \newcite{yuan-etal-2020-faithfulness} propose to optimize the factual consistency of e-commerce product summarization. 

%% file: conclusion_and_future_directions.tex
\section{Conclusion and Future Directions}

In this paper, we first introduce the \textit{factual inconsistency} problem in abstractive summarization. 
Then we provide an overview of approaches to evaluating and improving the factual consistency of summarization systems. 
Considering the landscape that we paint in this paper, we foresee the following directions:
\begin{enumerate}
    \item \textbf{Optimization for Extrinsic Errors.} The existing factual consistency optimizations methods mainly focus on intrinsic errors, while ignoring extrinsic errors. We argue that the main reason for the extrinsic factual inconsistency errors is that the current maximum likelihood estimation training strategy cannot explicitly model the hard constraints between the document and the summary (e.g., the entities and quotations must appear in the document). How to model these constraints into the summary generation process is a problem worth exploring. And reinforcement learning may be a feasible way to achieve this goal.
    \item \textbf{Paragraph-level Metrics.} Most evaluation works focus on calculating sentence-level factual consistency without considering the relationship between sentences. Paragraph-level evaluation is more challenging and valuable.
    \item \textbf{Factual Consistency in other Conditional Text Generation.} Besides the standard text summarization task, other conditional text generation tasks such as image captioning and visual storytelling also suffer factual inconsistency errors and cross-modal factual consistency is more challenging than single-text.
\end{enumerate} 
The above-mentioned research directions are by no means exhaustive and are to be considered as guidelines for researchers wishing to address the \textit{factual inconsistency} problem in abstractive summarization.

%% file: main.bbl
\begin{thebibliography}{}

\bibitem[\protect\citeauthoryear{Banko \bgroup \em et al.\egroup
  }{2007}]{10.5555/1625275.1625705}
Michele Banko, Michael~J. Cafarella, Stephen Soderland, Matt Broadhead, and
  Oren Etzioni.
\newblock Open information extraction from the web.
\newblock In {\em Proceedings of the 20th International Joint Conference on
  Artifical Intelligence}, IJCAI'07, page 2670–2676, San Francisco, CA, USA,
  2007. Morgan Kaufmann Publishers Inc.

\bibitem[\protect\citeauthoryear{Bordes \bgroup \em et al.\egroup
  }{2013}]{10.5555/2999792.2999923}
Antoine Bordes, Nicolas Usunier, Alberto Garcia-Dur\'{a}n, Jason Weston, and
  Oksana Yakhnenko.
\newblock Translating embeddings for modeling multi-relational data.
\newblock In {\em Proceedings of the 26th International Conference on Neural
  Information Processing Systems - Volume 2}, NIPS'13, page 2787–2795, Red
  Hook, NY, USA, 2013. Curran Associates Inc.

\bibitem[\protect\citeauthoryear{Cao \bgroup \em et al.\egroup }{2018}]{Cao:18}
Ziqiang Cao, Furu Wei, Wenjie Li, and Sujian Li.
\newblock Faithful to the original: Fact aware neural abstractive
  summarization.
\newblock In {\em Proceedings of the Thirty-Second {AAAI} Conference on
  Artificial Intelligence, (AAAI-18), the 30th innovative Applications of
  Artificial Intelligence (IAAI-18), and the 8th {AAAI} Symposium on
  Educational Advances in Artificial Intelligence (EAAI-18), New Orleans,
  Louisiana, USA, February 2-7, 2018}, pages 4784--4791, 2018.

\bibitem[\protect\citeauthoryear{Cao \bgroup \em et al.\egroup
  }{2020}]{cao-etal-2020-factual}
Meng Cao, Yue Dong, Jiapeng Wu, and Jackie Chi~Kit Cheung.
\newblock Factual error correction for abstractive summarization models.
\newblock In {\em Proceedings of the 2020 Conference on Empirical Methods in
  Natural Language Processing (EMNLP)}, pages 6251--6258, Online, November
  2020. Association for Computational Linguistics.

\bibitem[\protect\citeauthoryear{Devlin \bgroup \em et al.\egroup
  }{2019}]{devlin-etal-2019-bert}
Jacob Devlin, Ming-Wei Chang, Kenton Lee, and Kristina Toutanova.
\newblock {BERT}: Pre-training of deep bidirectional transformers for language
  understanding.
\newblock In {\em Proceedings of the 2019 Conference of the North {A}merican
  Chapter of the Association for Computational Linguistics: Human Language
  Technologies, Volume 1 (Long and Short Papers)}, pages 4171--4186,
  Minneapolis, Minnesota, June 2019. Association for Computational Linguistics.

\bibitem[\protect\citeauthoryear{Dong \bgroup \em et al.\egroup
  }{2020}]{dong-etal-2020-multi-fact}
Yue Dong, Shuohang Wang, Zhe Gan, Yu~Cheng, Jackie Chi~Kit Cheung, and Jingjing
  Liu.
\newblock Multi-fact correction in abstractive text summarization.
\newblock In {\em Proceedings of the 2020 Conference on Empirical Methods in
  Natural Language Processing (EMNLP)}, pages 9320--9331, Online, November
  2020. Association for Computational Linguistics.

\bibitem[\protect\citeauthoryear{Durmus \bgroup \em et al.\egroup
  }{2020}]{durmus-etal-2020-feqa}
Esin Durmus, He~He, and Mona Diab.
\newblock {FEQA}: A question answering evaluation framework for faithfulness
  assessment in abstractive summarization.
\newblock In {\em Proceedings of the 58th Annual Meeting of the Association for
  Computational Linguistics}, pages 5055--5070, Online, July 2020. Association
  for Computational Linguistics.

\bibitem[\protect\citeauthoryear{Falke \bgroup \em et al.\egroup
  }{2019}]{falke-etal-2019-ranking}
Tobias Falke, Leonardo F.~R. Ribeiro, Prasetya~Ajie Utama, Ido Dagan, and Iryna
  Gurevych.
\newblock Ranking generated summaries by correctness: An interesting but
  challenging application for natural language inference.
\newblock In {\em Proceedings of the 57th Annual Meeting of the Association for
  Computational Linguistics}, pages 2214--2220, Florence, Italy, July 2019.
  Association for Computational Linguistics.

\bibitem[\protect\citeauthoryear{Fan \bgroup \em et al.\egroup
  }{2018}]{fan-etal-2018-hierarchical}
Angela Fan, Mike Lewis, and Yann Dauphin.
\newblock Hierarchical neural story generation.
\newblock In {\em Proceedings of the 56th Annual Meeting of the Association for
  Computational Linguistics (Volume 1: Long Papers)}, July 2018.

\bibitem[\protect\citeauthoryear{Gabriel \bgroup \em et al.\egroup
  }{2020}]{gabriel2020figure}
Saadia Gabriel, Asli Celikyilmaz, Rahul Jha, Yejin Choi, and Jianfeng Gao.
\newblock Go figure! a meta evaluation of factuality in summarization, 2020.

\bibitem[\protect\citeauthoryear{Goodrich \bgroup \em et al.\egroup
  }{2019}]{10.1145/3292500.3330955}
Ben Goodrich, Vinay Rao, Mohammad Saleh, and Peter~J. Liu.
\newblock Assessing the factual accuracy of generated text.
\newblock {\em Proceedings of the 25th ACM SIGKDD International Conference on
  Knowledge Discovery and Data Mining}, 2019.

\bibitem[\protect\citeauthoryear{Gunel \bgroup \em et al.\egroup
  }{2019}]{gunel2019mind}
Beliz Gunel, Chenguang Zhu, Michael Zeng, and Xuedong Huang.
\newblock Mind the facts: Knowledge-boosted coherent abstractive text
  summarization.
\newblock In {\em NeurIPS 2019, Knowledge Representation and Reasoning Meets
  Machine Learning(KR2ML) Workshop}, December 2019.

\bibitem[\protect\citeauthoryear{Gupta}{2019}]{GUPTA201949}
Som Gupta.
\newblock Abstractive summarization: An overview of the state of the art.
\newblock {\em Expert Systems with Applications}, 121:49--65, 2019.

\bibitem[\protect\citeauthoryear{{Hahn} and {Mani}}{2000}]{881692}
U.~{Hahn} and I.~{Mani}.
\newblock The challenges of automatic summarization.
\newblock {\em Computer}, 2000.

\bibitem[\protect\citeauthoryear{Hermann \bgroup \em et al.\egroup
  }{2015}]{hermann2015teaching}
Karl~Moritz Hermann, Tomas Kocisky, Edward Grefenstette, Lasse Espeholt, Will
  Kay, Mustafa Suleyman, and Phil Blunsom.
\newblock Teaching machines to read and comprehend.
\newblock {\em NIPS}, 2015.

\bibitem[\protect\citeauthoryear{Huang \bgroup \em et al.\egroup
  }{2020}]{huang-etal-2020-knowledge}
Luyang Huang, Lingfei Wu, and Lu~Wang.
\newblock Knowledge graph-augmented abstractive summarization with
  semantic-driven cloze reward.
\newblock In {\em Proceedings of the 58th Annual Meeting of the Association for
  Computational Linguistics}, pages 5094--5107, Online, July 2020. Association
  for Computational Linguistics.

\bibitem[\protect\citeauthoryear{Koto \bgroup \em et al.\egroup
  }{2020}]{koto2020ffci}
Fajri Koto, Jey~Han Lau, and Timothy Baldwin.
\newblock Ffci: A framework for interpretable automatic evaluation of
  summarization, 2020.

\bibitem[\protect\citeauthoryear{Kryscinski \bgroup \em et al.\egroup
  }{2020}]{kryscinski-etal-2020-evaluating}
Wojciech Kryscinski, Bryan McCann, Caiming Xiong, and Richard Socher.
\newblock Evaluating the factual consistency of abstractive text summarization.
\newblock In {\em Proceedings of the 2020 Conference on Empirical Methods in
  Natural Language Processing (EMNLP)}, pages 9332--9346, Online, November
  2020. Association for Computational Linguistics.

\bibitem[\protect\citeauthoryear{Lewis \bgroup \em et al.\egroup
  }{2020}]{lewis-etal-2020-bart}
Mike Lewis, Yinhan Liu, Naman Goyal, Marjan Ghazvininejad, Abdelrahman Mohamed,
  Omer Levy, Veselin Stoyanov, and Luke Zettlemoyer.
\newblock {BART}: Denoising sequence-to-sequence pre-training for natural
  language generation, translation, and comprehension.
\newblock In {\em Proceedings of the 58th Annual Meeting of the Association for
  Computational Linguistics}, pages 7871--7880, Online, July 2020. Association
  for Computational Linguistics.

\bibitem[\protect\citeauthoryear{Li \bgroup \em et al.\egroup
  }{2018}]{li-etal-2018-ensure}
Haoran Li, Junnan Zhu, Jiajun Zhang, and Chengqing Zong.
\newblock Ensure the correctness of the summary: Incorporate entailment
  knowledge into abstractive sentence summarization.
\newblock In {\em Proceedings of the 27th International Conference on
  Computational Linguistics}, pages 1430--1441, Santa Fe, New Mexico, USA,
  August 2018. Association for Computational Linguistics.

\bibitem[\protect\citeauthoryear{Lin}{2004}]{lin-2004-rouge}
Chin-Yew Lin.
\newblock {ROUGE}: A package for automatic evaluation of summaries.
\newblock In {\em Text Summarization Branches Out}, pages 74--81, Barcelona,
  Spain, July 2004. Association for Computational Linguistics.

\bibitem[\protect\citeauthoryear{Mani and Maybury}{1999}]{mani1999advances}
Inderjeet Mani and T.~Mark Maybury.
\newblock Advances in automatic text summarization.
\newblock {\em Advances in Automatic Text Summarization}, 1999.

\bibitem[\protect\citeauthoryear{Mao \bgroup \em et al.\egroup
  }{2020}]{mao2020constrained}
Yuning Mao, Xiang Ren, Heng Ji, and Jiawei Han.
\newblock Constrained abstractive summarization: Preserving factual consistency
  with constrained generation, 2020.

\bibitem[\protect\citeauthoryear{Matsumaru \bgroup \em et al.\egroup
  }{2020}]{matsumaru-etal-2020-improving}
Kazuki Matsumaru, Sho Takase, and Naoaki Okazaki.
\newblock Improving truthfulness of headline generation.
\newblock In {\em Proceedings of the 58th Annual Meeting of the Association for
  Computational Linguistics}, pages 1335--1346, Online, July 2020. Association
  for Computational Linguistics.

\bibitem[\protect\citeauthoryear{Maynez \bgroup \em et al.\egroup
  }{2020}]{maynez-etal-2020-faithfulness}
Joshua Maynez, Shashi Narayan, Bernd Bohnet, and Ryan McDonald.
\newblock On faithfulness and factuality in abstractive summarization.
\newblock In {\em Proceedings of the 58th Annual Meeting of the Association for
  Computational Linguistics}, pages 1906--1919, Online, July 2020. Association
  for Computational Linguistics.

\bibitem[\protect\citeauthoryear{Mishra \bgroup \em et al.\egroup
  }{2020}]{mishra2020looking}
Anshuman Mishra, Dhruvesh Patel, Aparna Vijayakumar, Xiang Li, Pavan
  Kapanipathi, and Kartik Talamadupula.
\newblock Looking beyond sentence-level natural language inference for
  downstream tasks.
\newblock {\em arXiv preprint arXiv:2009.09099}, 2020.

\bibitem[\protect\citeauthoryear{Nallapati \bgroup \em et al.\egroup
  }{2016}]{Nallapati2016AbstractiveTS}
Ramesh Nallapati, Bowen Zhou, C.~D. Santos, Çaglar G{\"u}lçehre, and Bing
  Xiang.
\newblock Abstractive text summarization using sequence-to-sequence rnns and
  beyond.
\newblock In {\em CoNLL}, 2016.

\bibitem[\protect\citeauthoryear{Narayan \bgroup \em et al.\egroup
  }{2018}]{narayan-etal-2018-dont}
Shashi Narayan, Shay~B. Cohen, and Mirella Lapata.
\newblock Don{'}t give me the details, just the summary! topic-aware
  convolutional neural networks for extreme summarization.
\newblock In {\em Proceedings of the 2018 Conference on Empirical Methods in
  Natural Language Processing}, pages 1797--1807, Brussels, Belgium,
  October-November 2018. Association for Computational Linguistics.

\bibitem[\protect\citeauthoryear{Norouzi \bgroup \em et al.\egroup
  }{2016}]{norouzi2016reward}
Mohammad Norouzi, Samy Bengio, Navdeep Jaitly, Mike Schuster, Yonghui Wu, Dale
  Schuurmans, et~al.
\newblock Reward augmented maximum likelihood for neural structured prediction.
\newblock {\em Advances In Neural Information Processing Systems},
  29:1723--1731, 2016.

\bibitem[\protect\citeauthoryear{Papineni \bgroup \em et al.\egroup
  }{2002}]{papineni-etal-2002-bleu}
Kishore Papineni, Salim Roukos, Todd Ward, and Wei-Jing Zhu.
\newblock {B}leu: a method for automatic evaluation of machine translation.
\newblock In {\em ACL}, pages 311--318, 2002.

\bibitem[\protect\citeauthoryear{See \bgroup \em et al.\egroup
  }{2017}]{see-etal-2017-get}
Abigail See, Peter~J. Liu, and Christopher~D. Manning.
\newblock Get to the point: Summarization with pointer-generator networks.
\newblock In {\em Proceedings of the 55th Annual Meeting of the Association for
  Computational Linguistics (Volume 1: Long Papers)}, pages 1073--1083,
  Vancouver, Canada, July 2017. Association for Computational Linguistics.

\bibitem[\protect\citeauthoryear{Shah \bgroup \em et al.\egroup
  }{2021}]{shah2021nutribullets}
Darsh~J Shah, Lili Yu, Tao Lei, and Regina Barzilay.
\newblock Nutri-bullets: Summarizing health studies by composing segments,
  2021.

\bibitem[\protect\citeauthoryear{Vaswani \bgroup \em et al.\egroup
  }{2017}]{vaswani2017attention}
Ashish Vaswani, Noam Shazeer, Niki Parmar, Jakob Uszkoreit, Llion Jones,
  Aidan~N Gomez, {\L}ukasz Kaiser, and Illia Polosukhin.
\newblock Attention is all you need.
\newblock In {\em Advances in neural information processing systems}, pages
  5998--6008, 2017.

\bibitem[\protect\citeauthoryear{Wang \bgroup \em et al.\egroup
  }{2020}]{wang-etal-2020-asking}
Alex Wang, Kyunghyun Cho, and Mike Lewis.
\newblock Asking and answering questions to evaluate the factual consistency of
  summaries.
\newblock In {\em Proceedings of the 58th Annual Meeting of the Association for
  Computational Linguistics}, pages 5008--5020, Online, July 2020. Association
  for Computational Linguistics.

\bibitem[\protect\citeauthoryear{Yuan \bgroup \em et al.\egroup
  }{2020}]{yuan-etal-2020-faithfulness}
Peng Yuan, Haoran Li, Song Xu, Youzheng Wu, Xiaodong He, and Bowen Zhou.
\newblock On the faithfulness for {E}-commerce product summarization.
\newblock In {\em Proceedings of the 28th International Conference on
  Computational Linguistics}, pages 5712--5717, Barcelona, Spain (Online),
  December 2020. International Committee on Computational Linguistics.

\bibitem[\protect\citeauthoryear{Zajic \bgroup \em et al.\egroup
  }{2004}]{13730}
David Zajic, Bonnie~J Dorr, and R.~Schwartz.
\newblock Bbn/umd at duc-2004: Topiary.
\newblock {\em HLT-NAACL}, 2004.

\bibitem[\protect\citeauthoryear{Zhang* \bgroup \em et al.\egroup
  }{2020a}]{Zhang*2020BERTScore:}
Tianyi Zhang*, Varsha Kishore*, Felix Wu*, Kilian~Q. Weinberger, and Yoav
  Artzi.
\newblock Bertscore: Evaluating text generation with bert.
\newblock In {\em International Conference on Learning Representations}, 2020.

\bibitem[\protect\citeauthoryear{Zhang \bgroup \em et al.\egroup
  }{2020b}]{zhang-etal-2020-optimizing}
Yuhao Zhang, Derek Merck, Emily Tsai, Christopher~D. Manning, and Curtis
  Langlotz.
\newblock Optimizing the factual correctness of a summary: A study of
  summarizing radiology reports.
\newblock In {\em Proceedings of the 58th Annual Meeting of the Association for
  Computational Linguistics}, pages 5108--5120, Online, July 2020. Association
  for Computational Linguistics.

\bibitem[\protect\citeauthoryear{Zhao \bgroup \em et al.\egroup
  }{2020}]{zhao-etal-2020-reducing}
Zheng Zhao, Shay~B. Cohen, and Bonnie Webber.
\newblock Reducing quantity hallucinations in abstractive summarization.
\newblock In {\em Findings of the Association for Computational Linguistics:
  EMNLP 2020}, pages 2237--2249, Online, November 2020. Association for
  Computational Linguistics.

\bibitem[\protect\citeauthoryear{Zhou \bgroup \em et al.\egroup
  }{2020}]{zhou2020detecting}
Chunting Zhou, Jiatao Gu, Mona Diab, Paco Guzman, Luke Zettlemoyer, and Marjan
  Ghazvininejad.
\newblock Detecting hallucinated content in conditional neural sequence
  generation.
\newblock {\em arXiv preprint arXiv:2011.02593}, 2020.

\bibitem[\protect\citeauthoryear{Zhu \bgroup \em et al.\egroup
  }{2020}]{zhu2020boosting}
Chenguang Zhu, William Hinthorn, Ruochen Xu, Qingkai Zeng, Michael Zeng,
  Xuedong Huang, and Meng Jiang.
\newblock Boosting factual correctness of abstractive summarization with
  knowledge graph, 2020.

\end{thebibliography}
